\documentclass[11pt]{article}
\usepackage{arabtex}
\usepackage{coling2018}
\usepackage{utf8}
\setcode{utf8}
\usepackage{url}
\usepackage{graphicx} 
\usepackage{tikz}
\usepackage{caption} 
\usepackage[stable]{footmisc}
\date{}

\begin{document}

\title{\textbf{A Dataset of Kurdish (Sorani) Named Entities \\ {\small{\textit{An Amendment to Kurdish-BLARK Named Entities }}}}}

\author{
	\begin{tabular}[t]{c}
		Sazan Salar and	Hossein Hassani\\
		\textnormal{University of Kurdistan Hewl\^er}\\
		\textnormal{Kurdistan Region - Iraq}\\
		{\tt {\{sazan.salarnamiq, hosseinh}\}@ukh.edu.krd}
	\end{tabular}
}

\maketitle

\begin{abstract}

Named Entity Recognition (NER) is one of the essential applications of Natural Language Processing (NLP). It is also an instrument that plays a significant role in many other NLP applications, such as Machine Translation (MT), Information Retrieval (IR), and Part of Speech Tagging (POST). Kurdish is an under-resourced language from the NLP perspective. Particularly, in all the categories, the lack of NER resources hinders other aspects of Kurdish processing. In this work, we present a data set that covers several categories of NEs in Kurdish (Sorani). The dataset is a significant amendment to a previously developed dataset in the Kurdish BLARK (Basic Language Resource Kit). It covers 11 categories and 33261 entries in total. The dataset is publicly available for non-commercial use under CC BY-NC-SA 4.0 license at \url{https://kurdishblark.github.io/}.

\end{abstract}

\section{Introduction}
\label{sec:intro}

While working on Kurdish processing has been growing since \newcite{hassani2018blark} presented the status of resource availability for the language, it still needs more efforts to get the status of a ``computable''~\cite{hassani2018blark} language. One of the neglected or less-studied areas is Named Entity Recognition (NER), which has an important influence in many other language processing areas. Currently (as of December 2022), we could only find one work (\newcite{hassani2017method}) focusing on Kurdish NER, which is an introduction that addresses one particular aspect of NER, Proper Nouns.

We present a dataset of Proper Nouns in Kurdish. The dataset includes 33261 entries of 11 categories. It is publicly available for non-commercial use under CC BY-NC-SA 4.0 license at \url{https://kurdishblark.github.io/}. The next section presents a summary of the dataset.

\section{Dataset}
\label{sec:method}

We collected the data from various sources, namely, the Ministry of Higher Education [of Kurdistan Regional Government (KRG), Kurdistan Region of Iraq (KRI)], Chamber of Commerce (Erbil, KRI), Chamber of Commerce (Duhok, KRI), books (\newcite{ghfour2012}), and online resources. We extracted the names from \newcite{ghfour2012} and manually entered them into the dataset.  

Also, we transliterated some of the entries which were in Latin script into Persian-Arabic script. Finally, we preprocessed the lists to remove noises such as {\tiny{``\RL{زانكوى}''}} and {\tiny{``\RL{قوتابخانە}''}} form the beginning of the entities, and remove punctuation, such as commas and extra spaces, and eliminate duplicates.

Table~\ref{tab:nes} provides the summary of the dataset entries. 

\begin{table}[h]
	\begin{center}
		\caption{Kurdish Named Entity dataset - A summary of the entries.}
		\label{tab:nes}
		\scalebox{1}{
		\begin{tabular}{|c|c|}
			\hline
			Category &Number of Entries	\\ \hline \hline
			Person &10,955 \\ \hline
			Locations& 149 \\ \hline
			Villages &2643 \\ \hline
			Caves &69 \\ \hline
			Mountains& 198 \\ \hline
			Rivers &104 \\ \hline
			Universities& 35 \\ \hline
			Institutes& 13 \\ \hline
			Companies &12,316 \\ \hline
			Ministries &32 \\ \hline
			Schools &6747 \\ \hline
			\hline
			Total: & 33,261 \\ \hline
		\end{tabular}
	}
	\end{center}
\end{table}

\section*{Acknowledgment}
We appreciate various organizations and individuals for helping us during the data collection. Our special thank you extends to the Ministry of Education of the Kurdistan Regional Government (KRG) in the Kurdistan Region of Iraq, the Chambers of Commerce in Erbil and Duhok, and the Vision Education Office in KRI. We also appreciate Mr. Nechirvan Hussein for his assistance in collecting data.

\bibliographystyle{lrec}
\bibliography{KNERDataset}

\end{document}